\documentclass[runningheads]{llncs}
\usepackage[T1]{fontenc}
\usepackage{graphicx}
\usepackage{hyperref}

\usepackage{amsmath,amssymb,amsfonts}
\usepackage{ragged2e}
\usepackage{booktabs}
\usepackage{tikz}

\begin{document}
\title{MFP3D: Monocular Food Portion Estimation Leveraging 3D Point Clouds}
%
%
\author{Jinge Ma\inst{1} \and
Xiaoyan Zhang\inst{2} \and
Gautham Vinod\inst{1} \and
Siddeshwar Raghavan\inst{1} \and
Jiangpeng He\inst{1} \and
Fengqing Zhu\inst{1}}

\authorrunning{J. Ma et al.}
%
\institute{Elmore Family School of Electrical and Computer Engineering, Purdue University, West Lafayette, USA \and
College of Artificial Intelligence, Anhui University, Hefei, China}

\maketitle

\begin{abstract}
Food portion estimation is crucial for monitoring health and tracking dietary intake. Image-based dietary assessment, which involves analyzing eating occasion images using computer vision techniques, is increasingly replacing traditional methods such as 24-hour recalls. However, accurately estimating the nutritional content from images remains challenging due to the loss of 3D information when projecting to the 2D image plane. Existing portion estimation methods are challenging to deploy in real-world scenarios due to their reliance on specific requirements, such as physical reference objects, high-quality depth information, or multi-view images and videos. In this paper, we introduce MFP3D, a new framework for accurate food portion estimation using only a single monocular image. Specifically, MFP3D consists of three key modules: (1) a 3D Reconstruction Module that generates a 3D point cloud representation of the food from the 2D image, (2) a Feature Extraction Module that extracts and concatenates features from both the 3D point cloud and the 2D RGB image, and (3) a Portion Regression Module that employs a deep regression model to estimate the food's volume and energy content based on the extracted features. Our MFP3D is evaluated on MetaFood3D dataset, demonstrating its significant improvement in accurate portion estimation over existing methods. Code is available at  \href{https://github.com/jingema99/MFP3D.git}{https://github.com/jingema99/MFP3D.git}.

\end{abstract}
\section{Introduction}

The significance of a person's diet on their overall health and well-being is paramount. Chronic diseases such as diabetes are linked to poor dietary habits, therefore understanding one's nutritional intake is of utmost significance~\cite{liese2015dietary}. There has been a shift from traditional dietary methods towards image-based dietary assessment due to the ease of usage, fewer measurement or self-reporting errors, and improved accuracy in the estimation of nutritional content from eating occasion images~\cite{boushey2017new,poslusna2009misreporting}.  

However, accurate portion estimation is very challenging domain-specific problem compared to food recognition~\cite{min2023,mao2020_visual,he2023_icip,pan2023muti,he2023_long_tailed,he2021_iccv}. Even domain experts, such as trained dietitians, cannot accurately estimate the nutritional content of the food from eating occasion images alone~\cite{Shao2021EnergyDensity,he2020_mipr,he2021end}. Directly using monocular image for portion or nutrition estimation is an ill-posed problem due to the loss of 3D information when projecting from the 3D world coordinate to the 2D image plane. 

To combat this issue, many existing methods rely on various assumptions such as the availability of a physical reference in the image, such as a checkerboard pattern~\cite{vinod2024food}, or the presence of a high-quality depth map with real-world physical units of depth~\cite{thames2021nutrition5k}. Methods such as~\cite{Dehais2017TwoView,Konstantakopoulos2021StereoVision,lo2018food} rely on multiple views, videos, or depth maps, which may be difficult to obtain in real-world applications. Most existing methods that handle the 3D shape of food typically rely on input images with physical references~\cite{he2024metafood}, and few are able to solely depend on monocular images as input.

In this paper, we propose MFP3D, a new monocular food portion estimation pipeline, that reconstructs a point cloud representation of the food, and uses a multimodal approach for 3D and 2D feature adaptation for accurate portion estimation. Our MFP3D consists of three modules: 1) a \textit{3D Reconstruction Module} where the monocular image serves as the input to a depth-estimation network. The estimated depth map is then used to reconstruct a 3D point cloud representation of the food, 2) a \textit{Feature Extraction Module} which comprises a 3D feature extractor network and a 2D feature extractor network, and 3) a \textit{Portion Regression Module} where the extracted features are combined and passed through a deep regression model to estimate the food's volume and energy. Our MFP3D demonstrates significant improvements compared to existing methods on the MetaFood3D dataset~\cite{chen2024metafood3dlarge3dfood}, 
which includes 637 food objects across 108 categories, with diverse modalities and detailed nutritional data. 


The main contributions of our paper can be summarized as follows:
\begin{itemize}
    \item We introduce an end-to-end food portion estimation framework, which uses only a monocular RGB image as input and significantly outperforms existing methods without requiring additional information such as the depth map or physical references.  
    \item We have innovatively utilized 3D point cloud features for food portion estimation.
    \item We propose to combine the 2D image and corresponding 3D point cloud features in a multimodal approach for accurate portion estimation.
\end{itemize}

\section{Related Works}

\textbf{Food Portion Estimation.} Different classes of portion estimation methods use different representations or inputs to recreate the lost 3D information during image capture. 
These include multi-view methods~\cite{Dehais2017TwoView,Konstantakopoulos2021StereoVision}, depth-based methods~\cite{shao2023end,lo2018food}, model-based methods~\cite{vinod2024food}, and deep-learning based methods~\cite{Vinod2022DepthDomainAdaptation,thames2021nutrition5k}. The use of 3D food models in~\cite{vinod2024food} shows the efficacy of utilizing such representations of food. The method relies on using predefined 3D models of food and recreating the eating occasion image using the 3D model through object and camera pose estimation. 
However, the input to this method is constrained by the requirement of a physical reference (checkerboard pattern) in the eating occasion image. Further, food objects that don't fit the geometrical shape of its corresponding 3D model (e.g. whole avocado as compared to sliced avocado) will not achieve reasonable estimates for the food volume. Alternatively, the voxel reconstruction methods require some predefined knowledge of the scene such as in~\cite{thames2021nutrition5k} where the distance between the camera and image plane is a known constant. Further, the depth map captured in~\cite{thames2021nutrition5k} using a high-quality Intel RealSense RGBD camera makes it easy to capture the distance between the camera and the object in real-world units. However, without this information, there would need to be some scaling between the ground-truth volume and the voxel volume which would require knowledge of ground-truth volume for accurate results~\cite{vinod2024food}. Our method alleviates these concerns by reconstructing the 3D point cloud representation through an estimated depth map while also using its representation for portion estimation.



\noindent\textbf{3D Point Cloud.} 
3D point clouds can be sampled from real meshes obtained via 3D scanners or reconstructed from 2D images using existing methods such as depth estimation and 3D mesh reconstruction. Zoedepth\cite{bhat2023zoedepth} estimates depth maps for each pixel from monocular images, with depth values representing the coordinates of points in the third dimension. TripoSR\cite{tochilkin2024triposr} is one of the best-performing models for 3D mesh reconstruction from a single image. It directly reconstructs meshes, which can then be sampled to obtain 3D point clouds.

3D point cloud perception models extract features from a set of three-\linebreak dimensional coordinates, performing downstream tasks such as classification and segmentation. PointNet\cite{qi2017pointnet} was the first model introduced to handle unordered point cloud data. Improving upon its performance, CurveNet\cite{xiang2021walk} introduces continuous sequences of point segments, termed curves, into a ResNet-style network to enhance point cloud geometry learning by effectively aggregating features. Subsequent models introduced many improvements such as using advanced convolution, transformer structures, neighbor clustering, or various pre-training methods. While previous works focus on classification of 3D point clouds, we adapt a 3D point cloud feature extraction model for the regression of food portion. 

\section{Methodology}
\begin{figure*}[ht]
\centering
\includegraphics[width=\textwidth]{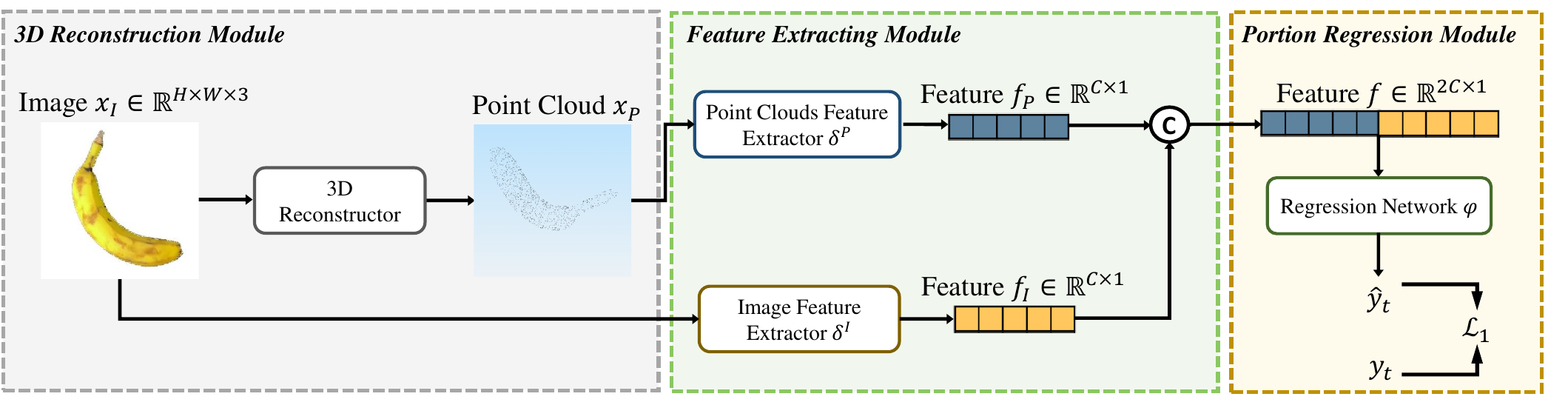} 
\caption{
An overview of the MFP3D framework: The input image $x_I$ goes through a three-stage pipeline for accurate portion estimation. In \textbf{Stage 1}, a 3D reconstructor is used to generate the point clouds from the input image. In \textbf{Stage 2}, the 3D features ($f_P$) of the point cloud and the 2D features ($f_I$) of the input image are extracted using networks $\delta_P$ and $\delta_I$, respectively. In \textbf{Stage 3}, these features are concatenated and passed through a regression network ($\varphi$) to estimate the food portion.}
\label{fig:pipeline}
\vspace{-10pt}
\end{figure*}

Our proposed MFP3D food portion estimation method derives fundamental quantitative attributes of food items such as shape, size, and texture from 3D point clouds and RGB images. The architecture of our three-stage pipeline is illustrated in Figure \ref{fig:pipeline}. In \textbf{Stage 1}, given an RGB image, $x\in {\mathbb{R}^{H\times W\times 3}}$, we first separate the each food item from the background using Segment Anything \cite{kirillov2023segment} to obtain the mask.  Next, we apply the mask to the original image, such that the processed image $x_I$ contains only the food. This processed image is then fed into a point cloud reconstruction model. This model generates a 3D representation $x_P$ of the food object from the single 2D image. In \textbf{Stage 2}, the image and its 3D representation are processed by two separate feature extractors: $\delta^I$ for the 2D image and $\delta^P$ for the 3D point clouds. These extractors produce feature maps $f_I$ and $f_P$, each with dimensions of $C \times 1$. The feature maps are then concatenated along the second axis to form a comprehensive feature vector $f \in \mathbb{R}^{2C \times 1}$. In \textbf{Stage 3}, the concatenated feature vector $f$ is fed into a deep regression module $\varphi$, which predicts the food portion $\hat{y}_t$. The attributes of $y_t$, such as energy content and volume, are defined by the ground truth labels used during training which are provided in the dataset. The pipeline is trained end-to-end in a supervised manner using the $\mathcal{L}1$ loss \cite{barron2019general}.



\subsection{3D Point Cloud Reconstruction}
\label{sec:PCrecon}

To effectively leverage 3D information, it is essential to acquire accurate 3D representations. In our study, point clouds are chosen as the preferred 3D format due to their lightweight storage requirements and their rich encapsulation of shape and size information. We explore four different types of point clouds to assess their impact on the performance of the portion estimation model. 

\begin{figure}[ht]
\vspace{-5pt}
\centering
\includegraphics[width=\columnwidth]{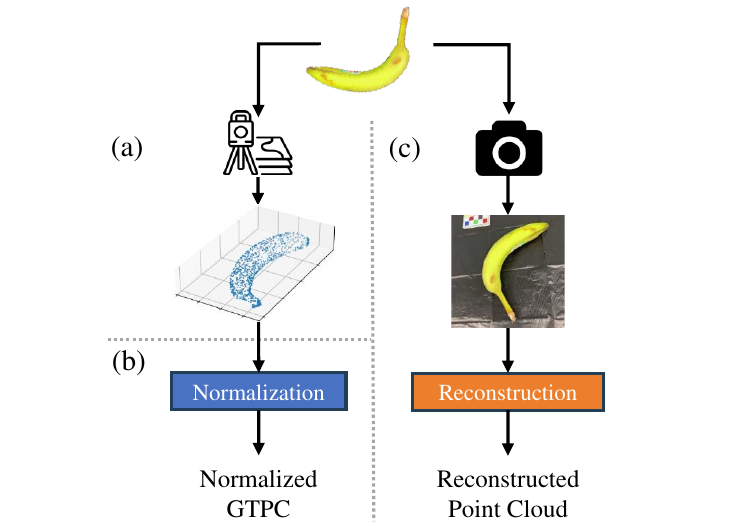} 
\caption{An overview of (a) Ground Truth Point Clouds (GTPC), (b) Normalized GTPCs and (c) Reconstructed Point Clouds, utilized in our experiments.}
\label{fig:PC}
\vspace{-10pt}
\end{figure}

\textbf{Ground Truth Point Clouds (GTPCs):} \label{subsec:GTPC}
GTPCs of food objects provide the most detailed and accurate representation of shape and size, enabling the network to achieve high precision in estimation results. We obtained these real point clouds by using a 3D scanner to capture the food items from multiple angles. From the original scans, we randomly sampled 1,024 points to derive the GTPCs, seen in Figure \ref{fig:PC}(a). In contrast, reconstructed point clouds may lose some of this detailed information, leading to less accurate results. Therefore, the performance of models based on GTPCs is considered the upper bound in our experiments.


The true scaling information of 3D point clouds is crucial for accurate portion estimation. However, current 3D reconstruction methods cannot obtain actual size reconstruction from monocular images, thus focusing only on shape. To fairly compare with methods that estimate portions solely from monocular images, as described in Figure \ref{fig:PC}(b), we \textbf{normalize GTPCs} to a range of $[0, 1]$, by rescaling all three dimensions of each point cloud to this range. This removes the true scaling information, allowing us to evaluate performance based on shape alone. 

\textbf{Reconstructed Point Clouds:}
Acquiring GTPCs requires specialized equipment, making it impractical for many applications. Therefore, we use point clouds reconstructed from monocular RGB images to simulate a more realistic scenario (as shown in Figure \ref{fig:PC}(c)). Any point cloud reconstruction model that accepts single images as input can be utilized. In our method, we adopt two types of generated point clouds: Depth point clouds and TripoSR point clouds.

For the depth point clouds, we use ZoeDepth \cite{bhat2023zoedepth} to estimate the depth map from a monocular image. Next, we segment the food foreground using masks from MetaFood3D, generated by Segment Anything \cite{kirillov2023segment}. To reconstruct the 3D point cloud, we retain the two original dimensions from the 2D image and incorporate the estimated depth as the third dimension. Finally, we randomly sample 1,024 points from the food foreground region to create the depth point cloud reconstruction.

For the TripoSR point clouds, we use the masks from MetaFood3D to generate images that retain only the food foreground. Then, we apply the TripoSR model \cite{tochilkin2024triposr}, which can directly reconstruct 3D meshes from monocular images and is widely used for this task. Finally, we randomly sample 1,024 points from the mesh to obtain the TripoSR point cloud reconstruction.


\subsection{Feature Extraction}
\label{sec:Feature}
The point cloud provides stereoscopic shape and size while the image includes ingredients, edges, and textures. Neither the point cloud modality nor the image modality alone can fully represent the complex information associated with food portion estimation. In this work, we propose to leverage information from both 2D and 3D representations to enhance the understanding of different aspects of the eating occasion image. By concatenating features extracted from the original 2D RGB image and the reconstructed 3D point cloud, our model can capture a more comprehensive view of the food object for better portion estimation.  

\textbf{2D Feature Extraction:}
We use an image feature extraction model $\delta^{I}(\cdot)$, built upon ResNet50~\cite{resnet} pre-trained on the ImageNet~\cite{imagenet} dataset. We exclude the last two layers of original ResNet50 but introduce an additional fully connected layer that maps the high-dimensional output to a lower-dimensional feature vector of length 512. This ensures a coherent and efficient feature representation. The overall feature extraction process can be formalized as:
\begin{equation}
    f_{I}^{i}=\delta^{I}(x_{I}^{i})
\end{equation}
where $f_{P}^{i}$ represents the 2D feature of the $i^{th}$ sample $x_{I}^{i}$.

\textbf{3D Feature Extraction:}
There exists many models designed for extracting features from point clouds. The pioneer network PointNet, known for its simplicity and efficiency, focuses on aggregating global features \cite{qi2017pointnet}. On the other hand, CurveNet’s ability to capture local details makes it superior for tasks requiring intricate local feature extraction \cite{xiang2021walk}. Therefore, CurveNet is selected as the backbone of the 3D feature extractor. The architecture of CurveNet consists of a Local Point Feature Aggregation (LPFA) module and a series of CurveNet Inception Convolutions (CIC). Firstly, LPFA aggregates local point features from the input point cloud, which is crucial for capturing fine-grained geometric details. Then CIC layers 
capture multi-scale features through point cloud down-sampling and feature extraction at various resolutions. After the CIC layers, convolutional and fully connected layers further process the aggregated features and map them to feature vector of the same size as the image features.
The 3D feature $f_{P}^{i}$ is formulated as:
\begin{equation}
    f_{P}^{i}=\delta^{P}(x_{P}^{i})
\end{equation}
where $\delta^{P}$ is the 3D feature extractor and $x_{P}^{i}$ is the reconstruction result of the $i^{th}$ sample $x_{I}^{i}$.

With features $f_{I}^{i}$ and $f_{P}^{i}$, we combine them together and form the comprehensive feature $f^{i}$. This is achieved by concatenating the two feature vectors, as follows:
\begin{equation}
    f^i=f_{I}^{i} \oplus f_{P}^{i}
\end{equation}
where $\oplus$ denotes the concatenation of the two vectors along the second axis. In this way, the integrated extractor has the strengths of both modalities by leveraging the geometric details from point clouds and the rich visual features from images.

\subsection{Portion Regression}
\label{sec:Portion}
For the portion estimation task, a numerical value is required to represent the final predictive result. To achieve this, we introduce a linear layer, denoted as $\varphi(\cdot)$, which maps the feature $f^{i}$ to a scalar value. By modifying the ground truth labels in the training data, the model can learn different parameter distributions based on the relationship between inputs and attributes. The model is defined as follows:
\begin{equation}
    \hat{y}^{i}_{t}=\varphi(f^{i})
\end{equation}
where $\hat{y}^{i}_{t}$ represents the estimated value of attribute $t$ for the $i^{th}$ sample.

For the loss function, we use L1 loss to measure the distance between the ground truths and the outputs. The L1 loss is given by:
\begin{equation}
    \mathcal{L}_1=\frac{1}{N^{\prime}}\sum_{i=1}^{N^{\prime}}{\left| \hat{y}_{t}^{i}-y_{t}^{i} \right|}
\end{equation}
where $y_{t}^{i}$ is the ground truth value for attribute $t$ of the $i^{th}$ sample and $N^{\prime}$ is the batch size.

\section{Experiments}

\subsection{Experimental setup}

\noindent \textbf{Dataset:} For our experiments, we utilize the publicly available dataset \linebreak MetaFood3D. This dataset includes 637 food objects across 108 categories. It is a comprehensive collection featuring 3D object meshes, 2D images, 3D point clouds, segmentation masks, RGBD video captures, nutritional information with weights, and blender renders with camera parameters for all the food items. We randomly select 510 food items for our training set, while 127 food items are \linebreak reserved for the test set. Since the MetaFood3D dataset is still under review, especially for base experiments, we also train and test our model on SimpleFood45~\cite{vinod2024food} for a more comprehensive evaluation.

\noindent \textbf{Implementation Details:} 
In the base experiments, we take a monocular food image as the input to 3D reconstruction module. It reconstructs a 3D point cloud from the image. The feature extracting module can extract food features solely from the point cloud, or jointly from both the point cloud and the image itself. We compared our method with various existing image-based energy estimation and volume estimation methods.

Our feature extraction network is designed to accommodate relatively flexible input data, such as point clouds reconstructed by different methods (or GTPC), or the option to use images as input. Therefore, in the ablation study, we compared the impact of using different point clouds on the model's performance, and also the effect of incorporating images as the input modality.

\noindent \textbf{Evaluation Metrics:} We employ two evaluation metrics to assess the precision of the model's estimation results. The first metric, Mean Absolute Error (MAE) \cite{willmott2005advantages}, calculates the average of the absolute errors in a set of predictions:
\begin{equation}
    \text{MAE} = \frac{1}{N}\sum_{i=1}^{N}{|\hat{y}_i - y_i|}
\end{equation}
where $\hat{y}_i$ is the prediction for the $i^{th}$ input, $y_i$ is the corresponding ground truth, and $N$ is the number of samples in the test batch. The second metric, Mean Absolute Percentage Error (MAPE) \cite{de2016mean}, expresses errors as a percentage, providing a clear depiction of the prediction error relative to the actual value:

\begin{equation}
    \text{MAPE} = \frac{100\%}{N}\sum_{i=1}^{N}{\frac{|\hat{y}_i - y_i|}{y_i}}
\end{equation}

\subsection{Experimental Results}
In this subsection, we compare our method MPF3D against existing image-based energy and volume estimation methods. We will also briefly introduce the key idea of each of the previous methods.

\noindent \textbf{Energy Estimation Methods:}
The \textit{baseline} model always predicts the mean volume and energy values from the dataset. The \textit{RGB only} approach utilizes a ResNet50 backbone and two linear layers to regress the energy estimates from an input image. The \textit{Density Map Only} method employs ground truth "Energy Density Maps"~\cite{Shao2021EnergyDensity} as input to regress the energy estimates. Instead of a regression network, the \textit{Density Map Summing} method sums up the values in the "Energy Density Maps" to estimate the energy. \textit{3D Assisted Portion Estimation} estimates both food volume and energy from 2D images using a physical reference in the eating scene.

Results are shown in Table~\ref{tab:energy} and Table~\ref{tab:SimpleFood45_energy}. By comparison, it can be observed that even without relying on the ground truth energy density map or physical reference as additional input or conditions, our method MPF3D still achieves the best results on both datasets, with the lowest MAE of 77.98 kCal and MAPE of 68.05\%.

\begin{table}[h!] 
\centering
\caption{Energy Estimation on MetaFood3D}
\scalebox{1.2}{ 
\begin{tabular}{lcc}
\toprule
Method & \multicolumn{2}{c}{Energy} \\
\cmidrule(lr){2-3} 
 & MAE(kCal)\(\downarrow\) & MAPE (\%)\(\downarrow\)\\
\midrule
Baseline  & 221.37 & 1,287.25\\
RGB Only \cite{Shao2021EnergyDensity} & 1,932.01 & 1,124.90\\
Density Map Only \cite{Shao2021EnergyDensity} & 1100.39 & 663.43 \\
Density Map Summing \cite{Ma2023DensityMapSumming} & 436.12 & 142.44\\
3D Assisted Portion Estimation \cite{vinod2024food}  & 260.79 & 102.25 \\
\textbf{MPF3D (Ours)} & \textbf{77.98} & \textbf{68.05}\\
\bottomrule
\end{tabular}
}
\label{tab:energy}
\end{table}

\begin{table}[h!] 
\centering
\caption{Energy Estimation on SimpleFood45}
\scalebox{1.1}{ 
\begin{tabular}{lcc}
\toprule
Method & \multicolumn{2}{c}{Energy} \\
\cmidrule(lr){2-3} 
 & MAE(kCal)\(\downarrow\) & MAPE (\%)\(\downarrow\)\\
\midrule
Baseline  & 120.09 & 547.34\\
RGB Only \cite{Shao2021EnergyDensity} & 273.56 & 222.72\\
Density Map Only \cite{Shao2021EnergyDensity} & 216.73 & 159.48 \\
Density Map Summing \cite{Ma2023DensityMapSumming} & 192.76 & 93.16\\
3D Assisted Portion Estimation \cite{vinod2024food}  &  32.01 & 25.13 \\
\textbf{MPF3D (Ours)} & \textbf{29.38} & \textbf{24.03}\\
\bottomrule
\end{tabular}
}
\label{tab:SimpleFood45_energy}
\end{table}

\noindent \textbf{Volume Estimation Methods:}
For volume estimation, we compare Stereo Reconstruction~\cite{Dehais2017TwoView}, Voxel Reconstruction~\cite{shao2023end}, baseline method against our MFP3D method as shown in Table~\ref{tab:volume} and Table~\ref{tab:SimpleFood45_volume} . The Voxel Reconstruction method~\cite{shao2023end} creates a voxel representation from the input image and corresponding depth maps, translating the number of occupied voxels into physical volume units. A regression network is trained to learn the relationship between voxel volume and ground truth volume, allowing for accurate volume estimation. Conversely, the Stereo Reconstruction method~\cite{Dehais2017TwoView} estimates food volume by capturing two images from different angles, using feature matching and triangulation to calculate depth. This depth information is used to reconstruct a 3D model of the food item, which is then analyzed to estimate the volume. 

Our method relies \textbf{solely on monocular images as the only input}, while other methods depend on additional information, such as binocular images, ground truth depth maps, or physical references. Through comparison, we found that our method can achieve performance close to or even surpassing other methods, despite using less information. On MetaFood3D, our method achieved the lowest MAE of 62.60 ml and MAPE of 41.43\%, while on SimpleFood45, our method performed comparably to Voxel Reconstruction and 3D Assisted Portion Estimation.

\begin{table}[h!] 
\centering
\caption{Volume Estimation on MetaFood3D}
\scalebox{1.2}{ 
\begin{tabular}{lcc}
\toprule
Method & \multicolumn{2}{c}{Volume} \\
\cmidrule(lr){2-3}
 & MAE(ml)\(\downarrow\)& MAPE(\%)\(\downarrow\)\\
\midrule
Baseline  & 151.85 & 845.69\\
Stereo Reconstruction \cite{Dehais2017TwoView}  & 135.96 & 210.90\\
Voxel Reconstruction  \cite{shao2023end} & 123.34 & 104.07\\
3D Assisted Portion Estimation \cite{vinod2024food}  &  195.92 & 79.33 \\
\textbf{MPF3D (Ours)} & \textbf{62.60} & \textbf{41.43}\\
\bottomrule
\end{tabular}
}
\label{tab:volume}
\end{table}

\begin{table}[h!] 
\centering
\caption{Volume Estimation on SimpleFood45}
\scalebox{1.2}{ 
\begin{tabular}{lcc}
\toprule
Method & \multicolumn{2}{c}{Volume} \\
\cmidrule(lr){2-3}
 & MAE(ml)\(\downarrow\)& MAPE(\%)\(\downarrow\)\\
\midrule
Baseline  & 83.28 &  170.37\\
Voxel Reconstruction  \cite{shao2023end} & \textbf{22.35} & 24.51\\  
3D Assisted Portion Estimation \cite{vinod2024food}  & 24.51 & \textbf{14.01} \\
\textbf{MPF3D (Ours)} & 25.83 & 16.15\\
\bottomrule
\end{tabular}
}
\label{tab:SimpleFood45_volume}
\end{table}

Our results indicate that the MFP3D method holds significant advantages over existing methods for energy and volume estimation. This is reflected in either a lower estimation error or a reduced requirement for input data.


\subsection{Ablation Studies}

In the ablation studies, we design a series of comparative experiments on Metafood3D to analyze: 
\begin{enumerate}
    \item The impact of using different 3D point clouds as input to the feature extraction module on the model's portion estimation performance.
    \item The effect of using RGB images as an additional input modality on the model's performance.
    \item The critical information within the point cloud for portion estimation.
\end{enumerate}

The various 3D  point clouds used include GTPC (as described in subsection~\ref{subsec:GTPC} and considered to be the upper bound), Normalized GTPC (without true scaling information), TripoSR~\cite{tochilkin2024triposr}, and Depth Point Clouds~\cite{bhat2023zoedepth}. It is worth noting that we used GTPC and Normalized GTPC only as control groups in the ablation studies. We did not use them in the base experiments because they can not be retrieved from monocular images but rather from 3D scanners.

We trained 8 different MFP3D models, as shown in Table ~\ref{tab:PCvsPCRGB}. 

\begin{table*}[ht] 
\centering
\caption{Ablation studies on different point clouds and the use of RGB images in \textbf{MPF3D}.}
\scalebox{1.02}{ 
\begin{tabular}{lcccc}
\toprule
Input to Feature Extraction & \multicolumn{2}{c}{Energy} & \multicolumn{2}{c}{Volume} \\
\cmidrule(lr){2-3} \cmidrule(lr){4-5}
 & MAE(kCal)\(\downarrow\) & MAPE (\%)\(\downarrow\) & MAE(ml)\(\downarrow\)& MAPE(\%)\(\downarrow\)\\
\midrule
\textbf{Point Cloud Only} & & & & \\
Upperbound - GTPCs & $\overline{114.73}$ & $\overline{71.00}$ & $\overline{26.06}$ & $\overline{19.19}$ \\
Normalized GTPCs & \textbf{135.61} & 114.62 & \textbf{79.93} & 68.05 \\
Depth Point Clouds \cite{bhat2023zoedepth}& 155.24 & \textbf{108.53} & 80.41 & \textbf{62.65} \\
TripoSR Point Clouds \cite{tochilkin2024triposr} & 175.45 & 152.02 & 121.80 & 83.47 \\
\multicolumn{5}{c}{\begin{tikzpicture}
\draw[thick, dashed] (0,0) -- (10,0);
\end{tikzpicture}} \\
\textbf{Point Cloud+RGB Image } & & & & \\
Upperbound - GTPCs & $\overline{26.16}$ & $\overline{17.37}$ {\fontsize{6}{10}\selectfont (-53.63)} & $\overline{26.68}$ & $\overline{15.59}$  {\fontsize{6}{10}\selectfont (-3.6)}\\
Normalized GTPCs & 100.96 & \textbf{62.65} {\fontsize{6}{10}\selectfont (-51.97)} & \textbf{49.26} & 42.19 {\fontsize{6}{10}\selectfont (-25.86)} \\
Depth Point Clouds & \textbf{77.98} & 68.05 {\fontsize{6}{10}\selectfont (-40.48)} & 62.60 & 41.43 {\fontsize{6}{10}\selectfont (-21.22)} \\
TripoSR Point Clouds & 109.64 & 98.45 {\fontsize{6}{10}\selectfont (-53.57)} & 62.41 & \textbf{39.45} {\fontsize{6}{10}\selectfont (-44.02)} \\ 
\bottomrule
\end{tabular}
}
\label{tab:PCvsPCRGB}
\end{table*}

The main differences between these MPF3D models lie in: (1) the type of 3D point cloud used, and (2) whether 2D RGB images are also used as input. The top half of Table ~\ref{tab:PCvsPCRGB} displays the model performance with portion estimate using only the 3D point cloud as input, while the bottom half shows the model performance when both the 3D point cloud and 2D RGB image are used as input, as illustrated in Figure~\ref{fig:pipeline}. In Table ~\ref{tab:PCvsPCRGB}, excluding the upperbound results from GTPC, the best result for each metric is bolded. In the bottom half of the table, we used small fonts to indicate the changes in MAPE for the models based on point cloud + RGB image compared to those based solely on the same point cloud.

\noindent \textbf{Observations}
\begin{enumerate}
    \item \textbf{Different point clouds:} We observed that GTPC achieved upper bound performance in both energy and volume estimation. Depth Point Clouds obtained the lowest Energy MAPE and volume MAPE among the point cloud-only methods. In the point cloud + RGB image methods, Depth Point Clouds achieved the lowest Energy MAE, while TripoSR obtained the lowest MAPE. We can infer that normalized GTPC does not offer a significant advantage over Depth Point Clouds and TripoSR Point Clouds extracted from monocular images.
    
    \item \textbf{Multimodality input:} We observed that adding RGB images as supplementary 2D input improved the performance of all models using the same point cloud across the board (as indicated by the small font in the table), though the degree of improvement varied. The percentage decrease in MAPE for volume estimation was less than that for energy estimation. For example, GTPC saw only a 3.6\% decrease in volume MAPE after adding RGB images, but a 53.63\% decrease in energy MAPE. We believe this may be because the point cloud data includes accurate volume information but lacks the food type, composition, and other energy-related information that might be present in RGB images. This suggests that incorporating multimodal information is crucial for accurate portion estimation.
    
    \item \textbf{Important information within the point cloud:} We observed that GTPC performed significantly better than other point clouds reconstructed from monocular images, but normalized GTPC did not show a clear advantage over the above methods. The difference between the two lies in the inclusion of the ground truth scaling factor. Therefore, we can infer that, in addition to the shape of the point cloud, the true scaling factor also contains critical information for portion estimation.
\end{enumerate}

\section*{Conclusion}
In this paper, we introduce MFP3D for estimating food portions by leveraging the combined power of 3D point clouds and 2D RGB images. This approach enhances the accuracy of volume and energy estimations and simplifies the data acquisition process by utilizing existing 3D point cloud reconstruction methods. These methods reduce dependency on difficult-to-obtain real-world 3D point cloud data and enable the reconstruction of point clouds from monocular images without additional annotations, providing superior performance and demonstrating the practical applicability of our approach. For future work, we plan to improve existing 3D reconstruction algorithms to obtain point clouds that more accurately represent the actual size of objects and explore additional data modalities such as textual descriptions and videos. Our results demonstrate that our method significantly improves energy and volume estimates, showcasing its great potential for real-world applications deployment.

%
%
%
%

\end{document}